\documentclass[journal,twoside,web]{ieeecolor}
\usepackage{jsen,jsen_compat,cite}
\usepackage{amsmath,amssymb,bm,booktabs,tabularx,array,multirow,graphicx}
\usepackage{tikz}
\usetikzlibrary{arrows.meta,positioning,fit,calc}
\usepackage{microtype}
\usepackage{placeins,wrapfig}
\usepackage{hyperref}
\hypersetup{hidelinks}
\newcolumntype{Y}{>{\centering\arraybackslash}X}
\definecolor{abstractbg}{rgb}{0.89804,0.94510,0.83137}

\begin{document}
\raggedbottom
\title{Polarity--Time Interaction and Multi-Level Reliability for Event-Based Spacecraft Pose Estimation}
\author{\mbox{Yonglong Zhang, Zongwu Xie, and Yang Liu\textsuperscript{\(\dagger\)}}%
\thanks{\textsuperscript{\(\dagger\)}Corresponding author: Yang Liu (e-mail: \href{mailto:liuyanghit@hit.edu.cn}{liuyanghit@hit.edu.cn}).}%
\thanks{All authors are with the State Key Laboratory of Robotics and Systems, Harbin Institute of Technology, Harbin 150001, Heilongjiang, China.}}

\IEEEtitleabstractindextext{%
\fcolorbox{abstractbg}{abstractbg}{%
\begin{minipage}{\textwidth}
\setlength{\intextsep}{0pt}
\setlength{\columnsep}{10pt}
\begin{wrapfigure}[15]{r}{0.34\textwidth}
\centering
\includegraphics[width=\linewidth]{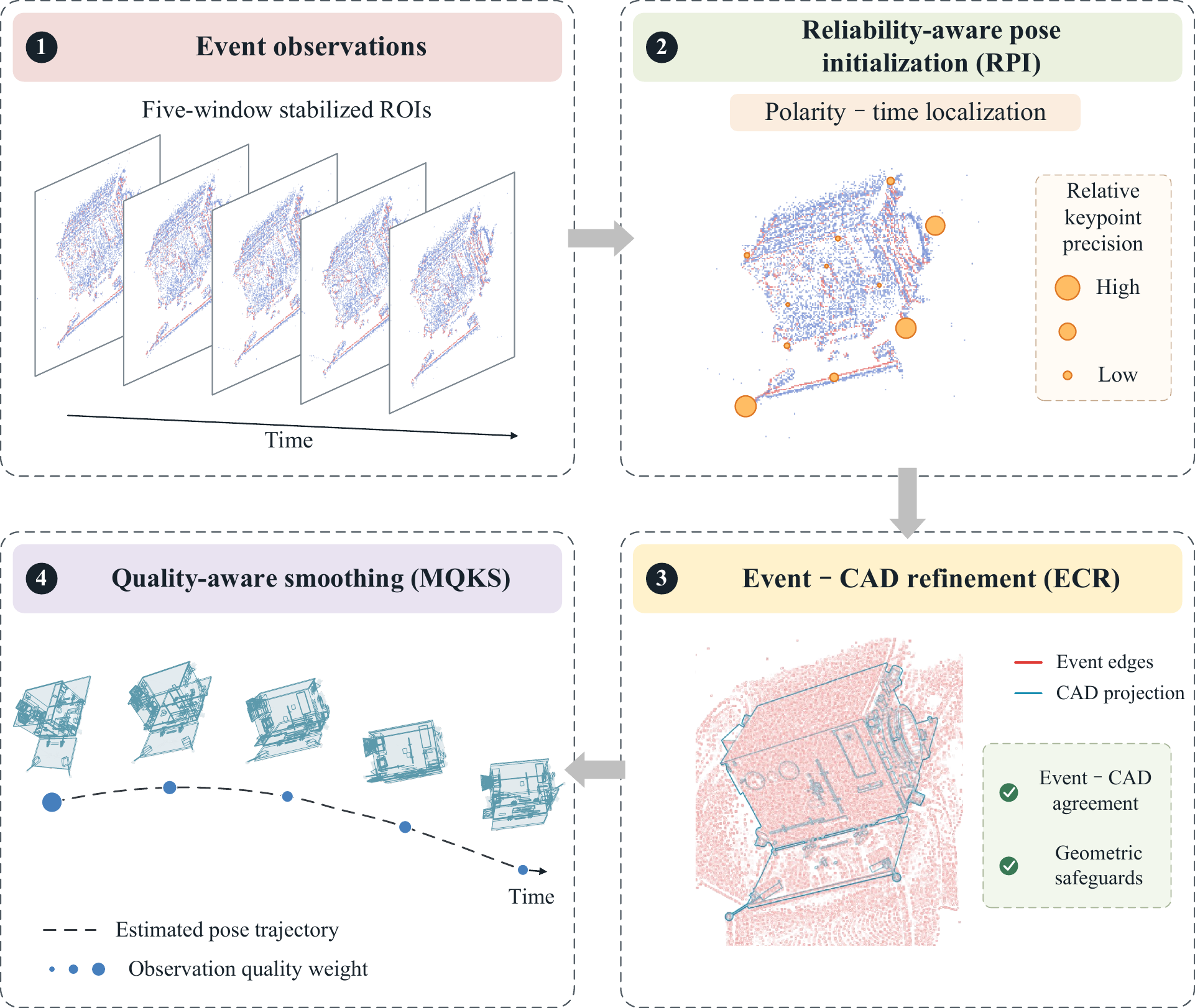}
\end{wrapfigure}
\begin{abstract}
Sparse event responses make spacecraft keypoints and successive pose observations unevenly reliable. The proposed event-based framework estimates reliability for both keypoint correspondences and accepted pose observations. A polarity--time interaction network processes positive counts, negative counts, and recent-event timing from five stabilized regions. The resulting local heatmap responses, together with ordered computer-aided design (CAD) geometry, predict relative point precision for robust pose recovery. Event--model refinement scores local candidates by event--CAD agreement, with geometric and sampling-consistency checks governing acceptance. The quality descriptor of the accepted observation then weights translation and rotation separately during trajectory optimization. On the official SPADES challenge test set, the complete system achieves a combined pose score of 0.0474, approximately 5.4\% below PASAN+IRIS. Validation ablations reduce the score from 0.0772 for base localization with uniform geometric weights to 0.0514 for the complete cascade. With identical per-frame observations, quality weighting reduces the score from 0.0549 under constant-weight optimization to 0.0514. These comparisons support estimating reliability at both correspondence and pose levels. The five-window configuration requires 52.7\,ms/frame under the reported evaluation setting.
\end{abstract}

\begin{IEEEkeywords}
Event-based vision, spacecraft pose estimation, keypoint reliability, geometric refinement, trajectory optimization
\end{IEEEkeywords}
\end{minipage}}}
\maketitle
\section{Introduction}

Noncooperative spacecraft pose estimation supplies the relative position and attitude needed for rendezvous, inspection, and on-orbit servicing. Shadows, reflections, and large changes in apparent target size complicate passive monocular measurements in space \cite{cassinis2019review,pauly2023survey}. Event cameras report the location, polarity, and time of brightness changes asynchronously \cite{lichtsteiner2008sensor,gallego2022event}. Their fine temporal resolution and dynamic range suit moving targets, but the spatial support of an event observation depends on motion and illumination. Sensor characterization also shows that event responses vary with operating conditions and settings \cite{delaney2025weather}. The resulting pose problem involves both locating the spacecraft and determining how much influence to assign to uneven measurements.

SPADES provides a dataset and evaluation protocol for event-based spacecraft pose estimation \cite{rathinam2024spades}. Polarity-aware localization and spatiotemporal aggregation have been combined with kinematic smoothing \cite{zeng2026spatiotemporal}; keypoint initialization has also been followed by event--CAD matching and trajectory optimization \cite{zhang2026cascaded}. Efficient onboard implementations address the computational requirements of event sensing \cite{rathinam2026onboard}. Within these pipelines, weights assigned during geometric recovery and after refinement determine each measurement's influence on the pose sequence.

For a keypoint correspondence, local heatmap shape describes response concentration. Its effect on pose also depends on the camera ray and the arrangement of the other points on the computer-aided design (CAD) model. A candidate pose selected from sparse event edges introduces a different decision: whether its event agreement justifies an update under the keypoint geometry and projection-sampling constraints. Once that update is accepted, the residuals and conditioning may change. Temporal weighting therefore needs to describe the accepted observation. The framework in this paper connects correspondence weighting to candidate acceptance and then to observation weighting.

Figure~\ref{fig:overall} shows target localization and region-of-interest (ROI) stabilization followed by reliability-aware pose initialization (RPI), event--CAD consistency refinement (ECR), and multi-source quality-aware kinematic smoothing (MQKS). The contributions concern how reliability is estimated and used across these stages:
\begin{enumerate}
\item Local heatmap evidence, camera rays, and context from 11 ordered CAD points (CAD11) are combined to predict relative point precision for geometric recovery. Training couples localization-error likelihood with a pose-error propagation proxy, accounting for the effect of correspondence errors on translation and rotation.
\item Event--CAD refinement uses unsigned orientation agreement and unit-mass overlap to score local poses. An update must satisfy geometric admissibility, improve on a same-pool geometric reference, and select the same candidate at two projection-integration resolutions.
\item A ten-dimensional quality descriptor is evaluated at the final per-frame pose. Heatmap shape, reprojection consistency, and geometric conditioning determine separate translation and rotation scales for trajectory optimization at the observations' physical timestamps.
\end{enumerate}

The experiments test these decisions under matched inputs: reliability learning uses fixed coordinates, and temporal weighting uses identical pose observations. Cascade ablations assess whether refinement gains persist after smoothing. The complete system achieves a combined score of 0.0474 on the official SPADES challenge test set and 0.0514 on the validation set.

\section{Related Work}

\subsection{Visual Spacecraft Pose Estimation}

Monocular spacecraft pose estimation includes direct regression and hybrid methods that recover pose from intermediate geometric measurements \cite{cassinis2019review,pauly2023survey}. Learned landmarks followed by nonlinear refinement \cite{chen2019landmark} and hierarchical estimation over a wide depth range \cite{hu2021wide} are examples of the hybrid route. Evaluation has progressed from the SPEED challenge \cite{kisantal2020challenge} to SPEED+, which includes synthetic images and hardware-in-the-loop observations \cite{park2022speedplus}. Domain transfer remains central in these settings: SPNv2 combines keypoint prediction, pose regression, and segmentation with online refinement \cite{park2023spnv2}, while recent work studies self-distillation for cross-domain spacecraft pose estimation from optical sensor data \cite{feng2025selfdistillation}.

In hybrid recovery, efficient perspective-n-point (EPnP) and random sample consensus (RANSAC) initialize pose and reject outliers \cite{lepetit2009epnp,fischler1981ransac}. Learning can also incorporate the geometric solution: BPnP differentiates through PnP \cite{chen2020bpnp}, and EPro-PnP learns weighted correspondences through a probabilistic pose objective \cite{chen2022epropnp}. The reliability head used here takes local responses together with fixed semantic CAD context. Its localization-error and propagated-pose-error objectives train the weights assigned to correspondences during robust recovery.

\subsection{Event-Based Pose Estimation and Refinement}

Event measurements can be reconstructed as intensity images for conventional vision algorithms \cite{rebecq2021highspeed} or represented through polarity and timing \cite{gallego2022event}. In spacecraft estimation, event sensing has been studied for synthetic-to-real transfer under illumination changes \cite{jawaid2023bridging}. Event--model alignment also supplies a criterion for selecting test-time adaptation examples in certifiable self-supervision \cite{jawaid2024certifiable}. SPADES provides evaluation on a realistic spacecraft target \cite{rathinam2024spades}; EDOPT addresses pose tracking of dynamic objects with event cameras \cite{glover2024edopt}.

The closest complete pipelines couple learned localization with geometry and temporal estimation. Zeng et al. use polarity-aware interaction, adjacent-window aggregation, PnP, and kinematic smoothing \cite{zeng2026spatiotemporal}. Zhang et al. add event--CAD matching between keypoint initialization and trajectory smoothing \cite{zhang2026cascaded}. Here, unsigned orientation moments and unit-mass overlap score pose candidates, while geometric admissibility, score gains, and agreement between integration levels govern the update itself. A geometric fallback supplies the observation when event evidence is insufficient.

\subsection{Reliability and Temporal Estimation}

Reliability enters geometric estimation through measurement covariance and residual weights. In spacecraft navigation, heatmap-derived covariance has been combined with filter innovations \cite{cassinis2023adaptive}; MAC-VO learns correspondence covariance for residual weighting in stereo visual odometry \cite{qiu2025macvo}. Temporal augmentation and dynamic-aware attention provide another way to exploit context in monocular odometry \cite{lin2026temporal}.

For the present cascade, point precision and pose quality have different inputs and roles. Point precision weights geometric recovery within a frame. Pose quality describes the accepted observation after refinement, including its updated residuals and conditioning, and controls its influence on the trajectory. ECR therefore precedes the computation of pose-dependent quality for MQKS.

\begin{figure*}[!t]
\centering
\includegraphics[width=\textwidth]{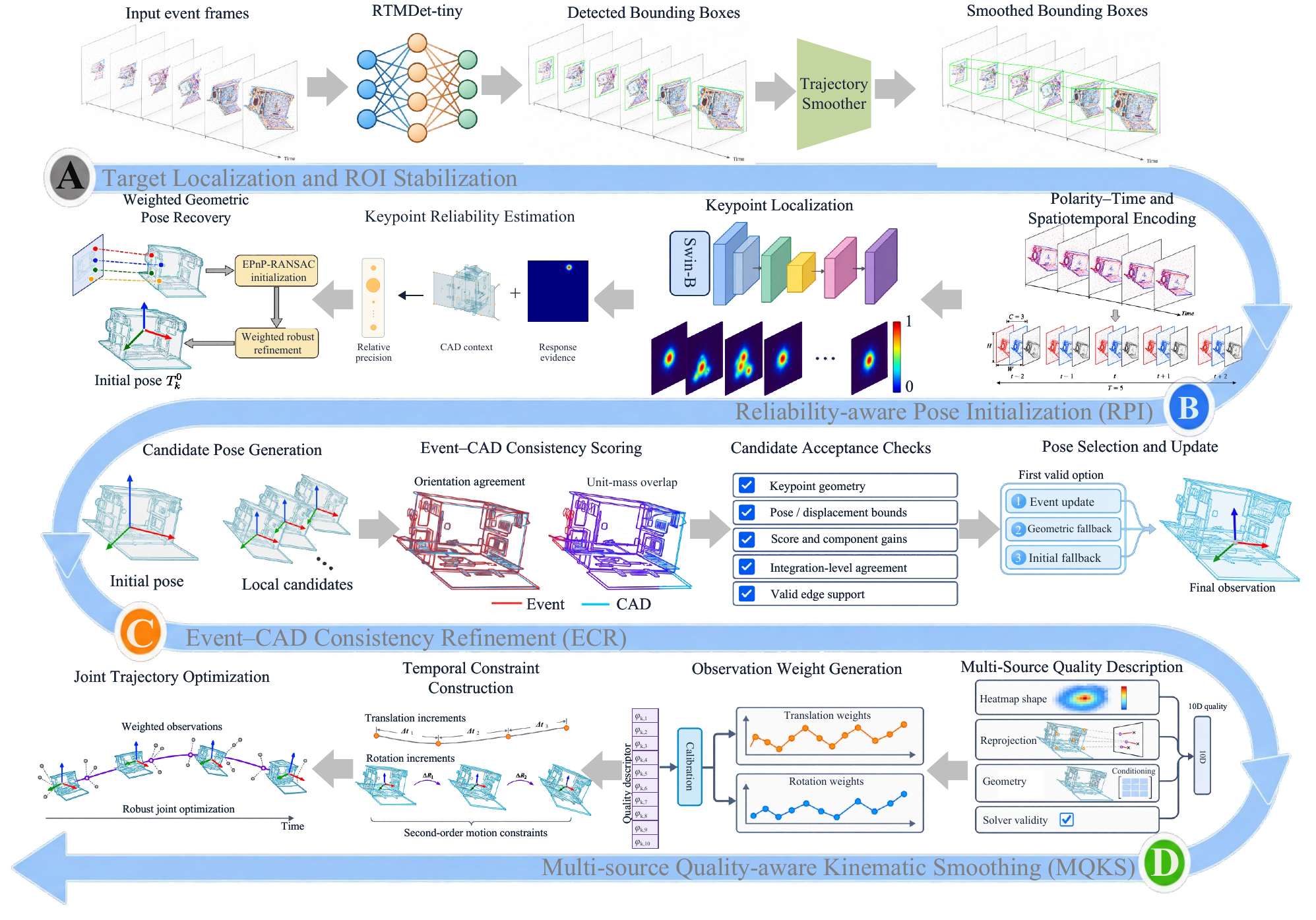}
\caption{Overall framework for event-based spacecraft pose estimation. (A) RTMDet-tiny detection and bounding-box trajectory smoothing define stabilized ROIs. (B) Polarity--time and spatiotemporal encoding supports semantic-keypoint localization; point reliability weights geometric recovery of the initial pose. (C) Local candidate generation, event--CAD consistency scoring, and acceptance checks produce the final per-frame observation through update or fallback. (D) Multi-source quality description and observation weighting are combined with temporal constraints for joint trajectory optimization. Follow the blue arrows through A--D; the gray arrows connect operations within each stage.}
\label{fig:overall}
\end{figure*}

\section{Method}\label{sec:method}

\subsection{Problem Definition and Overall Framework}

Given an event stream \(\mathcal E=\{(x_i,y_i,t_i,p_i)\}\), camera intrinsics \(K\), a spacecraft CAD mesh \(\mathcal M\), and 11 ordered CAD points \(\{\mathbf X_m\}_{m=1}^{11}\), the objective is to estimate the object-to-camera pose sequence

\begin{equation}
\label{eq:pose-projection}
\begin{aligned}
T_k&=(R_k,\mathbf t_k),\quad R_k\in SO(3),\\
\mathbf y_{k,m}&=R_k\mathbf X_m+\mathbf t_k,\\
\mathbf u_{k,m}&=\pi_K(\mathbf y_{k,m}).
\end{aligned}
\end{equation}

Here $k$ indexes output timestamps, $m\in\{1,\ldots,11\}$ indexes semantic CAD points, and $\pi_K$ is perspective projection with intrinsics $K$; $T\mathbf X$ abbreviates $R\mathbf X+\mathbf t$. We distinguish the RPI estimate $T_k^0$, the ECR observation $T_k^{\mathrm{obs}}$, and the final MQKS trajectory $\widehat T_k$. A per-frame observation refers to one output timestamp; its localization input still contains five event windows.

Stage A in Fig.~\ref{fig:overall} uses RTMDet-tiny \cite{lyu2022rtmdet} to detect the spacecraft in accumulated event frames. Local harmonic fitting smooths the box-center and size trajectories to suppress crop jitter from fluctuating event support \cite{zeng2026spatiotemporal}. The smoothed boxes define crops from the original event windows for keypoint localization.

At reference time \(t_k\), five adjacent event windows are mapped to a common ROI grid using the smoothed boxes. Their inverse transforms map predicted keypoints back to sensor coordinates. Polarity and recent-event timing provide within-window evidence, and the adjacent windows provide temporal context.

Stage B encodes the stabilized ROI sequence into 11 semantic heatmaps. Decoded coordinates establish 2D--3D correspondences for EPnP-RANSAC, followed by refinement with learned relative point precision to obtain \(T_k^0\).

Stage C tests local candidates against event--CAD evidence and geometry, yielding \(T_k^{\mathrm{obs}}\) through update or fallback. Stage D computes its quality and jointly optimizes the trajectory with separate translation and rotation weights and temporal constraints. Blue arrows in Fig.~\ref{fig:overall} indicate this order.

\begin{figure*}[!t]
\centering
\includegraphics[width=\textwidth]{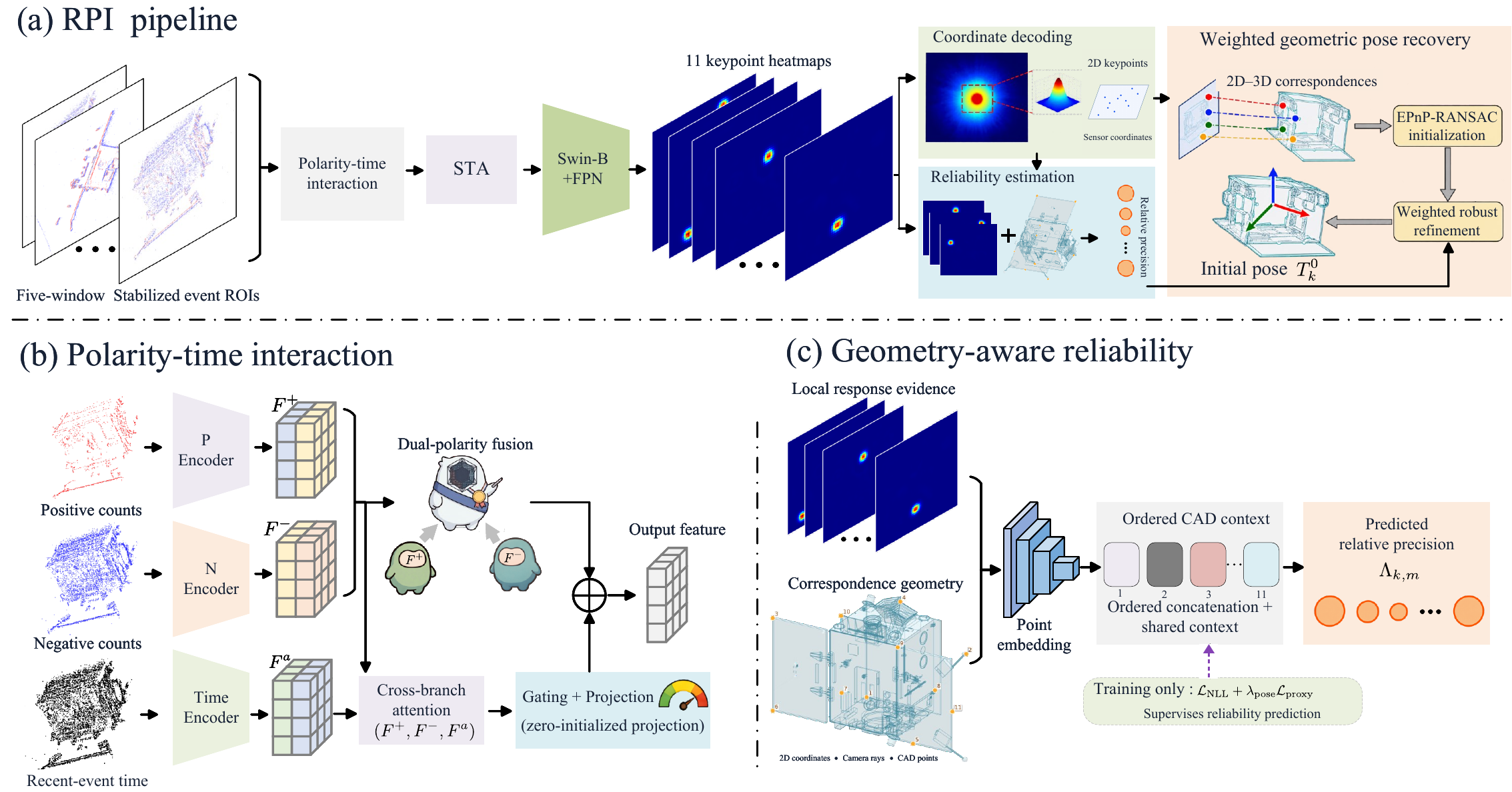}
\caption{Reliability-aware pose initialization (RPI). (a) Five-window encoding and localization produce 11 heatmaps. Decoded coordinates initialize EPnP-RANSAC; relative precision weights subsequent robust refinement. (b) Cross-branch attention combines positive counts, negative counts, and recent-event timing; gating and zero-initialized projection add a residual to dual-polarity features. (c) Local responses and correspondence geometry are encoded with ordered CAD context to predict relative precision, supervised by localization-error negative log-likelihood (NLL) and a pose-error propagation proxy during training only.}
\label{fig:rpi}
\end{figure*}

\subsection{Reliability-Aware Pose Initialization}

Figure~\ref{fig:rpi} details RPI: polarity--time encoding and spatiotemporal aggregation (STA) precede localization; the resulting heatmaps supply coordinates and relative precision for geometric recovery. Panel (c) expands reliability estimation.

\subsubsection{Polarity--Time and Spatiotemporal Encoding}

Polarity-aware attention and STA retain complementary polarity responses and aggregate neighboring windows \cite{zeng2026spatiotemporal}. Accumulated counts, however, assign the same value to different within-window event timings; cross-window aggregation cannot recover this discarded recency. We extend the dual-polarity representation with an explicit recent-event-time branch and three-way interaction before STA, allowing response recency to condition the polarity features used for localization.

For window \(W_j=[t_j^-,t_j^+)\), with $j\in\{k-2,\ldots,k+2\}$, let the positive- and negative-polarity counts at ROI pixel \(\mathbf u\) be \(C_j^+(\mathbf u)\) and \(C_j^-(\mathbf u)\). Fixed logarithmic compression produces \(\bar C_j^+\) and \(\bar C_j^-\), and the most recent event-time channel is defined as

\begin{equation}
\label{eq:event-time}
A_j(\mathbf u)=
\begin{cases}
\displaystyle\max_{i:\,t_i\in W_j,\ e_i\mapsto\mathbf u}
\displaystyle \frac{t_i-t_j^-}{t_j^+-t_j^-},&\text{if }\mathbf u\text{ receives events},\\
0,&\text{otherwise}.
\end{cases}
\end{equation}

Here $e_i=(x_i,y_i,t_i,p_i)$ and $e_i\mapsto\mathbf u$ denotes mapping the event to the cropped ROI. The per-window input is \(I_j=[\bar C_j^+,\bar C_j^-,A_j]\). The positive counts, negative counts, and recent-event time in Fig.~\ref{fig:rpi}(b) denote these three channels.

Three shallow encoders produce aligned feature maps \(F_j^+,F_j^-,F_j^a\). Within each window, channel cross-attention \cite{yan2025hvi} connects the three branches through six directed edges. For target branch \(b\) and source branch \(c\), queries are computed from the target, whereas keys and values are computed from the source:

\begin{equation}
\label{eq:cross-branch}
\begin{aligned}
M_{b\leftarrow c}
&=P^o_{b\leftarrow c}\!\left[
\operatorname{softmax}
(\tau_{b\leftarrow c}\widehat Q_b\widehat K_c^\top)V_c
\right],\\
\widetilde F_b&=F_b+\frac12\sum_{c\ne b}M_{b\leftarrow c}.
\end{aligned}
\end{equation}

In Eq.~\eqref{eq:cross-branch}, \(P^o\) is the output projection and \(\tau\) scales attention logits. Queries and keys are spatially normalized to model cross-channel correlations. All edges read pre-interaction features.

Channel-wise gates \(\alpha_b\) are predicted from the three global branch descriptors and normalized across branches, with $\sum_b\alpha_b=1$ for each channel, before constructing the residual:

\begin{equation}
\label{eq:gated-residual}
\begin{aligned}
F_j^{\mathrm{LCA}}
&=\operatorname{Concat}_{b\in\{+,-,a\}}
(3\alpha_b\odot\widetilde F_b),\\
F_j^{\mathrm{pol}}
&=\Phi_{\mathrm{dual}}([F_j^+,F_j^-])
+P_0(F_j^{\mathrm{LCA}}).
\end{aligned}
\end{equation}

Parameters are shared across five windows before STA alignment and aggregation. The zero-initialized \(P_0\) preserves the learned dual-polarity path \(\Phi_{\mathrm{dual}}\) at initialization, adding the three-branch interaction as a residual. Figure~\ref{fig:rpi}(b) omits the window index \(j\); \(F^a\) denotes the time-branch feature.

To align target structure across windows, STA predicts neighbor offsets relative to the center window and assigns temporal weights from paired descriptors:

\begin{equation}
\label{eq:sta}
\begin{aligned}
\widehat F_r
&=\operatorname{Warp}\!\left(
F_{k+r}^{\mathrm{pol}},
\Psi_{\mathrm{off}}([F_{k+r}^{\mathrm{pol}},F_k^{\mathrm{pol}}])\right),\quad r\ne0,\\
F_k^{\mathrm{STA}}
&=\rho\!\left([F_k^{\mathrm{pol}},
\sum_{r=-2}^{2}\beta_r\widehat F_r]\right).
\end{aligned}
\end{equation}

The center feature remains \(\widehat F_0=F_k^{\mathrm{pol}}\). Neighbor offsets \(\Psi_{\mathrm{off}}\) guide resampling by \(\operatorname{Warp}\), and \(\rho\) fuses the features. Paired source--center descriptors determine softmax-normalized weights \(\beta_r\). STA thus aggregates across windows after recency is encoded within each window.

\subsubsection{Keypoint Localization}

The aggregated feature $F_k^{\mathrm{STA}}$ is processed by Swin-B \cite{liu2021swin} and a feature pyramid network (FPN) \cite{lin2017fpn}, followed by a heatmap head that predicts $H_k=\{H_{k,m}\}_{m=1}^{11}$. Each semantic heatmap supplies both the coordinate decoder and the local response evidence for reliability estimation, as shown in Fig.~\ref{fig:rpi}(a). The coordinate decoder and ROI-to-sensor transforms are shared across the front-end ablations.

Each heatmap is decoded into continuous coordinates by a bounded local Gaussian decoder. Following distribution-aware localization \cite{zhang2020dark}, it fits a quadratic model to the log response around the dominant peak and falls back to the local centroid when curvature or offset checks fail. The inverse ROI transform maps each decoded point to the sensor-domain observation \(\widehat{\mathbf u}_{k,m}\in\mathbb R^2\). The heatmap peak gives selection confidence \(c_{k,m}\), which determines correspondence eligibility for geometric recovery.

\subsubsection{Keypoint Reliability Estimation}

Points admitted by a heatmap-confidence threshold can still differ in localization uncertainty and geometric sensitivity. We therefore learn their relative precision from local responses and ordered CAD geometry, so that weighting accounts for both response quality and the correspondence configuration. The reliability head in Fig.~\ref{fig:rpi}(c) uses the same heatmaps $H_k$ as coordinate decoding. For each point, a $5\times5$ patch around the peak is normalized by its maximum. The resulting 25 values are combined with two normalized ROI coordinates, three camera-ray components, three normalized CAD coordinates, response maximum, and local entropy to form a 35-dimensional descriptor.

A two-layer multilayer perceptron (MLP) embeds each point into 64 output channels. All 11 tokens are concatenated in fixed CAD order and projected to 128 channels to supply shared context for the positive scale readout $s_{k,m}$. The predicted scale is transferred from ROI to sensor coordinates before relative-precision normalization:

\begin{equation}
\label{eq:scale-transfer}
\Sigma_{k,m}^{\mathrm{ROI}}=s_{k,m}^2I_2,\qquad
\Sigma_{k,m}^{\mathrm{cam}}
=B_k\Sigma_{k,m}^{\mathrm{ROI}}B_k^\top,
\end{equation}

Here $I_2$ is the $2\times2$ identity and \(B_k\) is the linear part of the inverse ROI transform; \(\Sigma^{\mathrm{cam}}\) is a covariance in sensor pixels. Its inverse is divided by the mean half-trace over the selected set $\mathcal V_k$ defined below, then eigenvalue-clipped to $[0.1,4]$, yielding \(\Lambda_{k,m}\in\mathbb R^{2\times2}\). Under the isotropic scale and equal-axis ROI convention, \(\Lambda_{k,m}=\omega_{k,m}I_2\). Orange-circle sizes in Fig.~\ref{fig:rpi} indicate these relative weights. Thus $c_{k,m}$ selects correspondences, while $\Lambda_{k,m}$ determines their relative influence during refinement.

\subsubsection{Weighted Geometric Pose Recovery}

EPnP-RANSAC \cite{lepetit2009epnp,fischler1981ransac} initializes pose from \(\widehat{\mathbf u}_{k,m}\), CAD points \(\mathbf X_m\), and intrinsics \(K\). Refinement then minimizes

\begin{equation}
\label{eq:weighted-pose}
\begin{aligned}
T_k^0&=\arg\min_T\sum_{m\in\mathcal V_k}
\rho_p(\mathbf r_{k,m}^{\top}\Lambda_{k,m}\mathbf r_{k,m}),\\
\mathbf r_{k,m}&=\widehat{\mathbf u}_{k,m}-\pi_K(T\mathbf X_m).
\end{aligned}
\end{equation}

Here \(\mathcal V_k\) contains finite coordinates with \(c_{k,m}\ge0.3\), and \(\rho_p(z)=\delta_p^2\log(1+z/\delta_p^2)\) is the Cauchy penalty. The RPI output \(T_k^0\) denotes the pose after this weighted refinement and is passed to ECR.

The solver uses at most 16 local rotation/translation updates, Cauchy scale $\delta_p=4$ pixels, and diagonal damping $\kappa=10^{-4}$. Each update is bounded by 0.05 in normalized translation and 0.2 rad in rotation; backtracking tests factors $1,1/2,1/4,1/8$. With a zero-initialized scale head, the relative precision reduces exactly to the identity under the adopted ROI convention, reproducing the uniform-weight solver. Invalid initial estimates retain the established validity flags and fallback rules.

\begin{figure*}[!t]
\centering
\includegraphics[width=\textwidth]{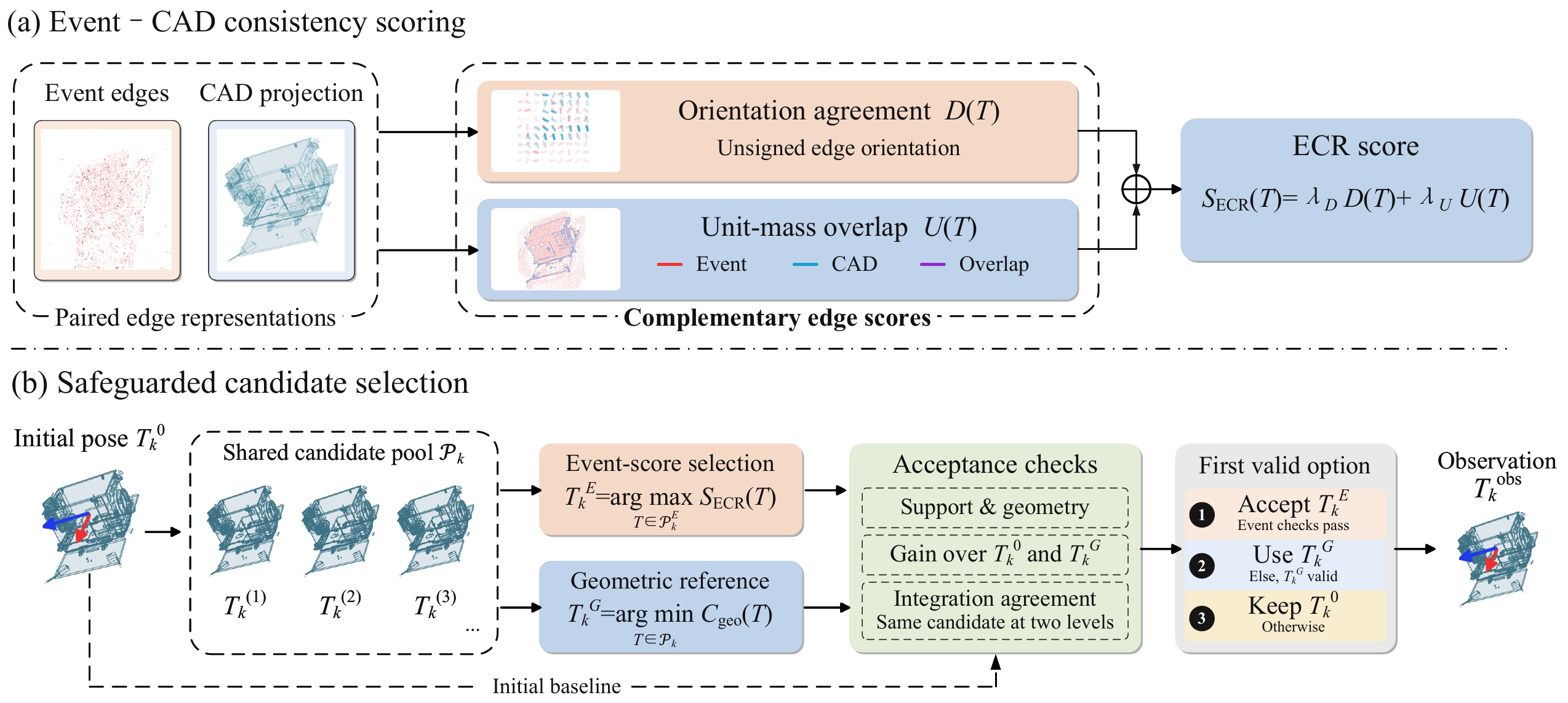}
\caption{Event--CAD consistency refinement (ECR). (a) Unsigned orientation agreement $D(T)$ and unit-mass overlap $U(T)$ form $S_{\mathrm{ECR}}(T)$. (b) A shared, geometrically admissible pool $\mathcal P_k$ provides the geometric reference $T_k^G$; event-score ranking uses its eligible subset $\mathcal P_k^E$. The check block summarizes support, geometry, gains over $T_k^0$ and $T_k^G$, and selection agreement across two integration levels. The first valid option yields $T_k^{\mathrm{obs}}$ for MQKS. CAD and edge glyphs visualize the corresponding representations.}
\label{fig:ecr-detail}
\end{figure*}

\subsubsection{Training Strategy and Objectives}

Training first adapts the polarity--time interaction with the inherited localization path fixed, then freezes localization to train reliability using the supervision in Fig.~\ref{fig:rpi}(c).

Localization adaptation uses ground-truth (GT) Gaussian heatmaps and teacher-heatmap retention, together with losses on inherited peak locations and confidence-based point selection:

\begin{equation}
\label{eq:front-loss}
\mathcal L_{\mathrm{front}}
=\mathcal L_{\mathrm{GT}}+\mathcal L_{\mathrm{KD}}
+\lambda_{\mathrm{peak}}\mathcal L_{\mathrm{peak}}
+\lambda_{\mathrm{entry}}\mathcal L_{\mathrm{entry}}.
\end{equation}

The first two terms of Eq.~\eqref{eq:front-loss} are mean-squared errors over valid keypoints against the GT Gaussian and the frozen teacher heatmap. The protection reference $H_m^0$ is obtained from the inherited localization path with the residual bypassed. Let $h_m(A)=\max_{\mathbf u\in A}\max(H_m(\mathbf u),0)$, and let $d_m(\mathbf u)$ be the sensor-space distance from a heatmap-cell center to the training GT projection. Define the peak-safe set $A_m$ by $d_m(\mathbf u)\leq e_m^0$, where $e_m^0$ is the decoded reference error; also include the inherited argmax and the grid cell nearest the GT projection. For the unit-peak GT Gaussian $G_m$, use the attainable margin
\begin{equation}
\label{eq:peak-margin}
\mu_m=\min\{[g_m(A_m)-g_m(A_m^c)]_+,
[h_m^0(A_m)-h_m^0(A_m^c)]_+\},
\end{equation}
where $g_m(A)=\max_{\mathbf u\in A}G_m(\mathbf u)$ and $[x]_+=\max(x,0)$. The peak loss is
\begin{equation}
\label{eq:peak-loss}
\mathcal L_{\mathrm{peak}}=
\frac1{N_v}\sum_{m\in\mathcal V}
[h_m(A_m^c)-h_m(A_m)+\mu_m]_+^2.
\end{equation}
With $\Omega$ denoting the heatmap domain, let $B_m=\{\mathbf u:d_m(\mathbf u)\leq8\ \mathrm{px}\}$ and $c_0=0.3$. Include the inherited peak in $B_m$ when its decoded error is within 8 px. With $a_m^0=\mathbb 1[h_m^0(\Omega)\geq c_0]$, the entry loss is
\begin{equation}
\label{eq:entry-loss}
\begin{split}
\mathcal L_{\mathrm{entry}}
={}&\frac1{N_v}\sum_{m\in\mathcal V}
\big\{a_m^0[c_0-h_m(\Omega)]_+^2\\
&+(1-a_m^0)\\
&\quad\cdot[h_m(B_m^c)-\max(h_m(B_m),c_0)]_+^2\big\}.
\end{split}
\end{equation}
Here $\mathcal V$ is the set of valid training points across the batch and $N_v=|\mathcal V|$; the frame index is suppressed in these training expressions. Empty-set maxima contribute no violation. These terms retain admitted reference points while permitting previously excluded points to enter through a geometrically accurate dominant response. All masks are detached from gradients and use GT only during training. During adaptation, the positive/negative encoders, dual-polarity fusion, STA, and localization tail are frozen; only the recent-event-time encoder, cross-branch attention, gates, and residual projection are updated. Gradients still propagate through the frozen downstream operations. The teacher is used only during training, and inference requires a single localization network.

Localization-error likelihood relates predicted scale to coordinate error, whose effect on pose also depends on the projection geometry. We add a local pose-error propagation proxy to train the relative weights through their effect on translation and rotation. Reliability training holds the adapted localization outputs, coordinates, and local geometry fixed; only the reliability head is optimized:

\begin{equation}
\label{eq:reliability-loss}
\mathcal L_{\mathrm{rel}}
=\mathcal L_{\mathrm{NLL}}
+\lambda_{\mathrm{pose}}\mathcal L_{\mathrm{proxy}}.
\end{equation}

Suppressing the frame index, let $\mathbf e_m$ be the ROI coordinate error and $\Sigma_m=\Sigma_{k,m}^{\mathrm{ROI}}$. The likelihood is
\begin{equation}
\label{eq:nll}
\mathcal L_{\mathrm{NLL}}=\frac1{2N_v}\sum_{m\in\mathcal V}
\left(\mathbf e_m^\top\Sigma_m^{-1}\mathbf e_m+
\log\det\Sigma_m+2\log(2\pi)\right).
\end{equation}
For the pose proxy, fix a valid predicted pose $\widetilde T$ and its confidence-selected support. Let $J_m$ be the projection Jacobian with respect to additive camera translation and left rotation, and let $D=\operatorname{diag}(\max(\|\widetilde{\mathbf t}\|,1)I_3,I_3)$ normalize translation. Define $\bar J_m=J_mD$, the predicted reprojection residual $\mathbf r_m$, and the training target $\mathbf d_m=\mathbf u_m^*-\widehat{\mathbf u}_m$. With relative precision $\Lambda_m$,
\begin{equation}
\label{eq:pose-proxy-system}
\begin{aligned}
W_m&=\frac{\Lambda_m}{1+\mathbf r_m^\top\Lambda_m\mathbf r_m/\delta_p^2},\\
A&=\sum_m\bar J_m^\top W_m\bar J_m,\qquad
\mathbf b=\sum_m\bar J_m^\top W_m\mathbf d_m,\\
\delta\boldsymbol\xi&=
\left[A+\kappa\operatorname{diag}\!\left(\max(\operatorname{diag}A,10^{-8})\right)\right]^{-1}\mathbf b.
\end{aligned}
\end{equation}
The proxy penalizes the propagated, dimensionless translation and rotation components:
\begin{equation}
\label{eq:pose-proxy-loss}
\mathcal L_{\mathrm{proxy}}=\operatorname{mean}_{k\in\mathcal V_f}
\left(\frac{\|\delta\boldsymbol\xi_{k,t}\|^2}{c_t}
+\frac{\|\delta\boldsymbol\xi_{k,R}\|^2}{c_R}\right).
\end{equation}
Here $\mathcal V_f$ indexes valid training frames, and $\theta$ denotes reliability-head parameters. The scales $c_t,c_R$ are mean squared propagated errors under identity precision on 256 fixed training examples. On that same batch, $\lambda_{\mathrm{pose}}=\|\nabla_\theta\mathcal L_{\mathrm{NLL}}\|/\|\nabla_\theta\mathcal L_{\mathrm{proxy}}\|$ is computed once and frozen. Jacobians and support are prediction-derived; GT supplies $\mathbf d_m$. The linear system in Eq.~\eqref{eq:pose-proxy-system} and the loss in Eq.~\eqref{eq:pose-proxy-loss} train the reliability head; inference uses Eq.~\eqref{eq:weighted-pose}. The fixed-coordinate comparisons in Table~\ref{tab:frontend-design} show the additional reduction in pose error from learned precision and the propagation proxy.

\subsection{Event--CAD Consistency Refinement}

ECR updates $T_k^0$ using the event--CAD score in Fig.~\ref{fig:ecr-detail}(a) and the safeguarded selection in panel (b), yielding the observation $T_k^{\mathrm{obs}}$ for MQKS.

\subsubsection{Candidate Pose Generation}

For a valid RPI estimate, local pose perturbations around $T_k^0$, including the zero perturbation, generate candidates. After the fixed keypoint-geometry and pose-displacement bounds are applied, the retained poses form the shared pool $\mathcal P_k$. Chronological and reverse-time event surfaces guide proposal generation; final acceptance uses the chronological observation. Both event and geometric selection draw from this pool, which is rebuilt at each RPI output. All candidates use a common spatial mask and rendering convention.

\subsubsection{Event--CAD Consistency Scoring}

Gradient-direction agreement and Soft-IoU provide complementary measures for event--model matching \cite{zhang2026cascaded}. A signed gradient comparison changes under contrast reversal, while direct strength overlap depends on the relative edge magnitudes. These quantities differ between brightness-change observations and geometric renderings. We use unsigned orientation moments to remove edge-sign ambiguity and unit-mass overlap to compare spatial edge distributions independently of their total strength. The observation map uses the EROS sequential surface \cite{gava2022puck}; visible CAD faces are rendered with a fixed normal-based intensity proxy. Both use the same evaluation grid and an initial-pose-dependent mask held fixed across candidates.

For either image, let \(\mathbf g\) be its Sobel gradient, \(a=\min(\|\mathbf g\|,1)\) the clipped magnitude, and \(\mathbf n\) the unit direction at nonzero-gradient locations. Magnitude and orientation moments are computed over the same neighborhood:

\begin{equation}
\label{eq:edge-moments}
\begin{aligned}
S&=\mathcal G_{\sigma_e}*a,\qquad
J=\mathcal G_{\sigma_e}*(a\mathbf n\mathbf n^\top),\\
\mathbf q&=\frac{(J_{11}-J_{22},\,2J_{12})^\top}
{\operatorname{tr}J},\qquad c=\|\mathbf q\|.
\end{aligned}
\end{equation}

Locations with zero support are assigned zero. The orientation moment and magnitude share the same support, so \(\operatorname{tr}J=S\). The double-angle representation \(\mathbf q\) is invariant to direction reversal, and \(c\) measures local orientation coherence. Event--CAD orientation affinity is

\begin{equation}
\label{eq:orientation-affinity}
A_{EC}=\tfrac12(c_Ec_C+\mathbf q_E^\top\mathbf q_C).
\end{equation}

Let $M_k$ be the fixed initial-pose mask. Applying it to the smoothed strengths gives $a_E=M_kS_E$ and $a_C=M_kS_C(T)$, with $p_E=a_E/\sum_{\mathbf u}a_E$ and $p_C=a_C/\sum_{\mathbf u}a_C$ on supported images. Subscripts $E$ and $C$ denote event and CAD fields; the CAD quantities depend on the candidate pose $T$. Orientation and distribution agreement are

\begin{equation}
\label{eq:ecr-score}
\begin{aligned}
D(T)&=\frac{\sum_{\mathbf u}a_Ea_CA_{EC}}
{\sqrt{(\sum_{\mathbf u}a_E^2)(\sum_{\mathbf u}a_C^2)}},\\
U(T)&=\frac{\sum_{\mathbf u}\min(p_E,p_C)}
{\sum_{\mathbf u}\max(p_E,p_C)},\\
S_{\mathrm{ECR}}(T)&=\lambda_DD(T)+\lambda_UU(T).
\end{aligned}
\end{equation}

The fixed weights \(\lambda_D,\lambda_U\) combine orientation and unit-mass overlap in Eq.~\eqref{eq:ecr-score}, as illustrated in Fig.~\ref{fig:ecr-detail}(a). Unsigned moments preserve agreement under opposite edge signs. The rendered CAD projection is area-integrated at the pixel level before gradients are computed, reducing sensitivity to boundary sampling. Event-based acceptance is disabled when valid magnitude or coherent-orientation support is absent. A common pair of area-integration levels is selected from reference convergence before event candidates are ranked.

\subsubsection{Candidate Acceptance Checks}

Event-score maximization alone imposes neither keypoint-consistency bounds nor agreement across rendering resolutions. We couple event gains to a geometric reference from the same candidate pool and test the selected update at two integration levels. The shared pool $\mathcal P_k$ contains candidates that satisfy bounds on robust reprojection cost, projected displacement, pose increments, and conditioning. Its geometric reference is
\begin{equation}
\label{eq:geometric-reference}
T_k^G=\arg\min_{T\in\mathcal P_k}C_{\mathrm{geo}}(T),
\end{equation}
where $C_{\mathrm{geo}}$ is the mean robust cost of the initially trusted keypoints. All candidates retain the same point support; ties with the initial pose retain $T_k^0$. Event eligibility further requires valid edge support, the proposal-stage admissibility checks, gains over both $T_k^0$ and $T_k^G$, and nondecreasing orientation and overlap terms. These conditions define $\mathcal P_k^E\subseteq\mathcal P_k$ before event-score ranking. Integration-level agreement is tested on the selected candidate.

\subsubsection{Pose Selection and Update}

At each of the two common integration levels, the event candidate is selected as
\begin{equation}
\label{eq:event-candidate}
T_k^E=\arg\max_{T\in\mathcal P_k^E}S_{\mathrm{ECR}}(T).
\end{equation}
The level pair is determined by convergence of $T_k^0$ and $T_k^G$ and held fixed for all event candidates. An event update requires a nonempty eligible set and the same winning candidate at both levels. If reference convergence or selection agreement fails, the geometric fallback is used. Figure~\ref{fig:ecr-detail}(b) summarizes the prioritized output:

\begin{equation}
\label{eq:candidate-selection}
T_k^{\mathrm{obs}}=
\begin{cases}
T_k^E,&\text{if all event and geometric tests pass},\\
T_k^G,&\text{if no event update passes and }T_k^G\text{ is valid},\\
T_k^0,&\text{otherwise}.
\end{cases}
\end{equation}

Invalid initial observations retain their validity flags. The geometric reference makes each accepted event update conditional on improvement beyond keypoint-only selection.

\subsection{Multi-Source Quality-Aware Kinematic Smoothing}

\subsubsection{Multi-Source Quality Description}

Kinematic constraints reduce temporal fluctuations, but the influence of each observation also depends on its weight. Heatmap-only quality remains unchanged when ECR updates a pose, even though its reprojection residuals and geometric conditioning can change. Stage D in Fig.~\ref{fig:overall} therefore evaluates quality at the accepted ECR pose and calibrates translation and rotation weights separately. These weights adapt observation influence to the refined geometry within joint trajectory optimization.

The quality vector contains geometric scales and conditioning, reprojection statistics, heatmap shape, and solver validity:

\begin{equation}
\label{eq:quality-descriptor}
\boldsymbol\phi_k=
[\boldsymbol\phi_k^{\mathrm{geo}},
\boldsymbol\phi_k^{\mathrm{rep}},
\boldsymbol\phi_k^{\mathrm{heat}},v_k]\in\mathbb R^{10}.
\end{equation}

The geometric descriptor uses the projection Jacobian and pixel covariances obtained from local heatmap second moments. The translation-normalized information matrix is stabilized, inverted, and inflated according to reprojection residuals to obtain translation and rotation scales. These pose-level scales come from heatmap moments; RPI learns point-level task scales.

After an ECR update, the reprojection and geometry terms are recomputed at the accepted pose. The four heatmap-shape terms remain fixed. Calibration and inference use the same ten-feature definition.

The three geometric entries are the log translation scale, log rotation scale, and log condition number of the normalized information matrix. Each scale is the square root of one third of its covariance-block trace; the condition number is clipped to $[1,10^{12}]$ before taking its logarithm. Quality computation retains points with confidence at least 0.3 and reprojection error at most 8 px, falling back to confidence-only support when fewer than four points remain. The two reprojection entries are $\log(1+\operatorname{median}_{m\in\mathcal S_k}\|\mathbf r_m\|)$ and $|\mathcal S_k|/11$, where $\mathcal S_k$ denotes this retained set.

The four heatmap entries are the mean peak response, normalized local entropy, secondary-to-primary peak ratio, and local-to-total response mass. Local statistics use a clipped $9\times9$ neighborhood around each peak and are averaged over the semantic points; entropy is divided by the logarithm of the neighborhood size. Peak responses and secondary ratios are clipped to $[0,1]$. A binary solver-validity indicator completes the descriptor.

\begin{figure*}[!t]
\centering
\includegraphics[width=.93\textwidth]{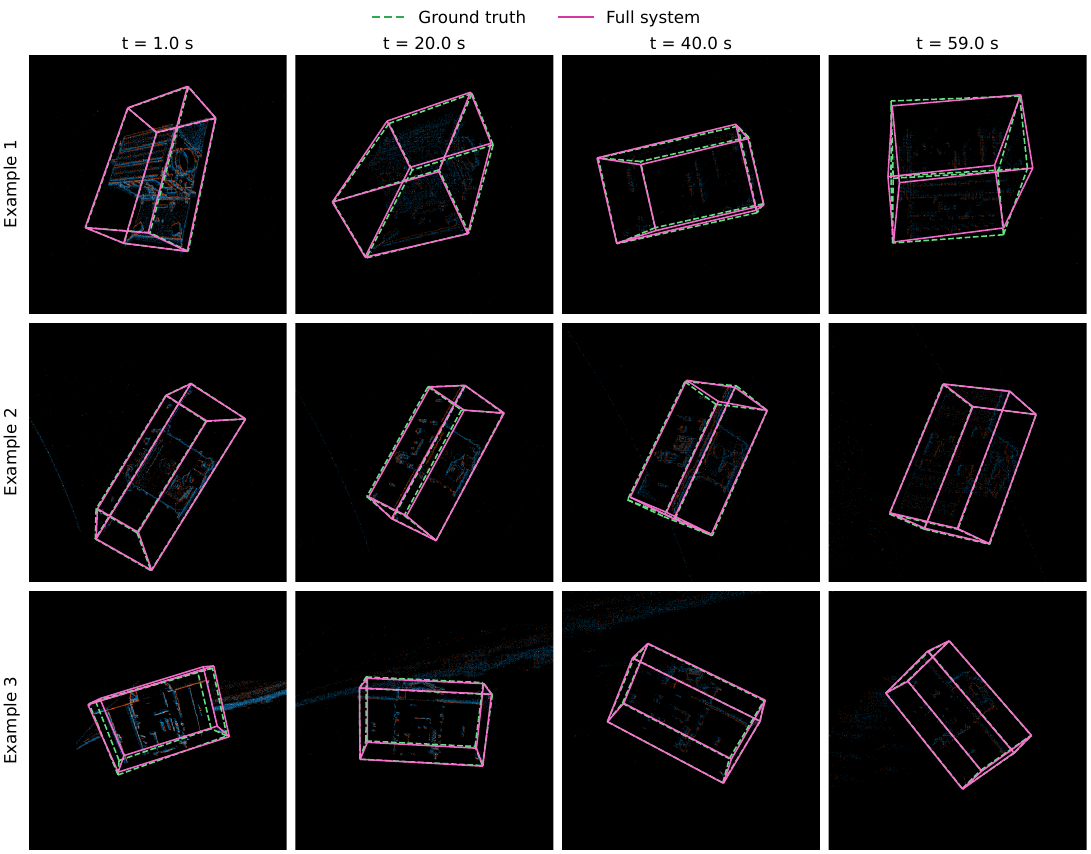}
\caption{Full-system estimates $\widehat T_k$ after MQKS on three preselected validation sequences, using seed 2026. Rows denote sequences and columns denote fixed timestamps. Green dashed and magenta solid lines show the CAD bounding box projected with the ground-truth and final estimated poses, respectively. Positive and negative events are orange and blue. All panels use a $448\times448$ display grid covering 1.5 times the original ROI side length.}
\label{fig:pose-overlays}
\end{figure*}

\subsubsection{Observation Weight Generation}

Translation and rotation error scales are calibrated separately from absolute translation errors in meters and rotation angles in radians. Let \(\widetilde{\boldsymbol\phi}_k\) be the standardized, clipped feature vector augmented with an intercept. Log-domain ridge regression on squared errors followed by radial scale correction gives

\begin{equation}
\label{eq:quality-scale}
\widehat\sigma_{k,c}
=\sqrt{\gamma_c
\exp(\boldsymbol\beta_c^\top\widetilde{\boldsymbol\phi}_k)},
\qquad c\in\{t,R\}.
\end{equation}

Regression and radial correction use calibration data. Standardized feature inputs and predicted scales are clipped to fixed bounds; at runtime, quality is computed from observations alone. Using the median calibration scale \(\sigma_c^{\mathrm{anc}}\) as an anchor yields the relative observation weight

\begin{equation}
\label{eq:observation-weight}
r_{k,c}=
\operatorname{clip}\!\left[
(\widehat\sigma_{k,c}/\sigma_c^{\mathrm{anc}})^\eta,
r_{\min},r_{\max}\right],\qquad
w_{k,c}=v_kr_{k,c}^{-2}.
\end{equation}

The fixed exponent \(\eta\) tempers relative scales, bounded by \(r_{\min},r_{\max}\). The frame-level weights $w_{k,t}$ and $w_{k,R}$ enter Eq.~\eqref{eq:observation-residual} at the corresponding timestamp. They are computed from post-ECR quality and held fixed during trajectory optimization; the point-level precisions $\Lambda_{k,m}$ are used in RPI. Table~\ref{tab:quality-sources} compares identical post-ECR observations and motion constraints, isolating the gain from these quality weights over constant weighting.

\subsubsection{Temporal Constraint Construction}

We use \(\operatorname{Log}(R)=(\log R)^\vee\in\mathbb R^3\) and translation increments \(\Delta\mathbf t_k=\mathbf t_{k+1}-\mathbf t_k\). Trajectory states are parameterized by additive translation and left-multiplicative rotation increments about the final per-frame observations. The numerical start uses Gaussian-smoothed translations and continuous relative rotation vectors with a four-frame standard deviation. Observation factors retain the individual measurements, while motion factors constrain changes in consecutive translation and rotation increments:

\begin{equation}
\label{eq:observation-residual}
\mathbf r_k^{\mathrm{obs}}=
\begin{bmatrix}
\sqrt{w_{k,t}}(\mathbf t_k-\mathbf t_k^{\mathrm{obs}})/s_t\\
\sqrt{w_{k,R}}\operatorname{Log}(R_k(R_k^{\mathrm{obs}})^{-1})/s_R
\end{bmatrix},
\end{equation}

\begin{equation}
\label{eq:motion-residual}
\Delta R_k=R_{k+1}R_k^{-1},\qquad
\mathbf r_k^{\mathrm{mot}}=
\begin{bmatrix}
(\mathbf t_{k+2}-2\mathbf t_{k+1}+\mathbf t_k)/m_t\\
\operatorname{Log}(\Delta R_{k+1}\Delta R_k^{-1})/m_R
\end{bmatrix}.
\end{equation}

\subsubsection{Joint Trajectory Optimization}

MQKS obtains the trajectory \(\widehat T_k=(\widehat R_k,\widehat{\mathbf t}_k)\) by jointly adjusting pose states under the weighted observation and second-order motion factors:

\begin{equation}
\label{eq:trajectory-objective}
\min_{\{R_k,\mathbf t_k\}}
\sum_k\rho_h(\|\mathbf r_k^{\mathrm{obs}}\|_2)
+\sum_k\rho_h(\|\mathbf r_k^{\mathrm{mot}}\|_2)
.
\end{equation}

Fixed scales \(s_t,s_R\) normalize observation residuals and \(m_t,m_R\) normalize motion residuals; \(\rho_h\) applies a Huber penalty to each six-dimensional block. Invalid observations receive zero weight and are interpolated for initialization; their states remain constrained by motion factors. Sparse nonlinear least squares retains the original, equally spaced grid, including invalid timestamps. Motion scales correspond to second-order differences at this fixed interval.

\section{Experiments and Results}

\subsection{Experimental Setup}

\subsubsection{Dataset and Evaluation Protocol} Experiments are organized on SPADES, which provides event observations of the Proba-2 spacecraft, camera calibration, a CAD model, and pose annotations \cite{rathinam2024spades}. The dataset description in Zhang et al. includes 300 synthetic trajectories containing 179,400 frames and 31 laboratory trajectories containing 15,500 frames \cite{zhang2026cascaded}. The proposed method uses events as its visual input and known camera--CAD geometry for pose recovery.

The evaluation separates validation ablations from the final test comparison. All proposed-method ablations use one predefined validation partition, with training and quality-calibration data kept separate from evaluation trajectories. The final checkpoint and inference settings are fixed before test evaluation. Table~\ref{tab:official-comparison} reports the complete-system test result alongside the published challenge scores \cite{zhang2026cascaded,zeng2026spatiotemporal}. The validation studies isolate cascade stages, front-end design, observation weighting, and input-window length.

\subsubsection{Evaluation Metrics} Translation and rotation errors are
\begin{equation}
\label{eq:evaluation-errors}
\begin{aligned}
e_{t,k}&=\frac{\|\widehat{\mathbf t}_k-\mathbf t_k^*\|_2}{\|\mathbf t_k^*\|_2},\\
e_{R,k}&=2\arccos\!\left(\operatorname{clip}(|\langle\widehat q_k,q_k^*\rangle|,0,1)\right).
\end{aligned}
\end{equation}
Here hats denote the output of the configuration being evaluated, stars denote ground truth, and $\widehat q_k,q_k^*$ are unit quaternions. The challenge metric sets translation errors below 0.002173 and rotation errors below $0.169^\circ$ to zero. Frame-averaged translation and rotation scores are denoted by $E_t$ and $E_R$, with rotation expressed in radians; the combined score is $E_{\mathrm{pose}}=E_t+E_R$. Lower scores indicate better performance. Pose-recovery failures are retained under the fixed evaluation policy. External baseline scores are taken from the cited studies.

\subsubsection{Implementation and Comparison Settings} The default configuration uses five input windows and 11 semantic CAD keypoints. Localization, reliability learning, ECR, and MQKS follow Section~\ref{sec:method}.

Internal comparisons use matched data, model-selection rules, and training budgets. Front-end variants are trained with their respective architectures; ECR comparisons share the same RPI outputs. Temporal comparisons use complete sequences, with quality mappings fitted for each observation stream by the same calibration procedure and frozen during evaluation. External methods retain their published pipelines. Table~\ref{tab:window-length} reports per-frame runtime for each window configuration, including 52.7\,ms/frame for five windows. Because the centered input requires future windows and MQKS optimizes complete sequences, the per-frame processing cost does not measure online latency.

\subsection{Comparison with Existing Methods}

The complete RPI--ECR--MQKS system has the lowest combined score among the methods listed in Table~\ref{tab:official-comparison} for the official SPADES challenge test set. PASAN+IRIS combines polarity-aware temporal localization with kinematic smoothing; the RF-DETR cascade uses keypoint initialization, photometric refinement, and trajectory smoothing \cite{zeng2026spatiotemporal,zhang2026cascaded}. Other entries retain their leaderboard team names. The proposed system obtains $E_t=0.0138$, $E_R=0.0336$, and $E_{\mathrm{pose}}=0.0474$, a combined reduction of approximately 5.4\% relative to PASAN+IRIS, with lower translation and rotation scores. The RF-DETR cascade retains the lower rotation score of 0.0331.

\begin{table}[!t]
\centering
\caption{Comparison on the official SPADES challenge test set \cite{zeng2026spatiotemporal,zhang2026cascaded}. Lower scores indicate better performance.}
\label{tab:official-comparison}
\small
\setlength{\tabcolsep}{4pt}
\begin{tabularx}{\columnwidth}{Xrrr}
\toprule
Method / reported team & $E_t$ & $E_R$ & $E_{\mathrm{pose}}$ \\
\midrule
PASAN+IRIS & 0.0146 & 0.0355 & 0.0501 \\
RF-DETR cascade & 0.0230 & \textbf{0.0331} & 0.0561 \\
haoranhuang & 0.1464 & 0.5562 & 0.7026 \\
pruthj & 0.1335 & 0.9101 & 1.0436 \\
csu nuaa pang & 0.3340 & 0.7724 & 1.1064 \\
zhaoyutao22 & 0.1959 & 1.0607 & 1.2565 \\
Kashyap Hegde Kota & 0.1121 & 1.1767 & 1.2888 \\
\midrule
Proposed complete system & \textbf{0.0138} & 0.0336 & \textbf{0.0474} \\
\bottomrule
\end{tabularx}
\end{table}

\subsection{Ablation Studies}

\subsubsection{Contributions of the Cascade Stages}

Table~\ref{tab:cascade-stages} starts from the dual-polarity network with STA and uniform geometric weights, matching the first row of Table~\ref{tab:frontend-design}. RPI adds polarity--time localization and correspondence reliability. The remaining comparisons distinguish per-frame structural refinement by ECR from temporal optimization by MQKS: RPI+MQKS omits refinement, while the complete cascade tests whether the ECR gain persists in the final trajectory. ECR-enabled and ECR-disabled configurations share their RPI predictions, and MQKS is evaluated over complete continuous sequences.

RPI lowers the combined score from 0.0772 to 0.0697, and ECR further reduces it to 0.0674. The temporal comparison tests whether this per-frame gain survives smoothing: RPI+MQKS gives 0.0529, compared with 0.0514 for the complete cascade. The latter score of 0.0514 is approximately 33.4\% below the base configuration.

\begin{table}[!t]
\centering
\caption{Stage-wise ablation on the validation set.}
\label{tab:cascade-stages}
\small
\begin{tabularx}{\columnwidth}{Xrrr}
\toprule
Configuration & $E_t$ & $E_R$ & $E_{\mathrm{pose}}$ \\
\midrule
Base localization + uniform PnP & 0.0208 & 0.0564 & 0.0772 \\
RPI & 0.0191 & 0.0506 & 0.0697 \\
RPI + ECR & 0.0185 & 0.0489 & 0.0674 \\
RPI + MQKS & 0.0154 & 0.0375 & 0.0529 \\
RPI + ECR + MQKS & \textbf{0.0150} & \textbf{0.0364} & \textbf{0.0514} \\
\bottomrule
\end{tabularx}
\end{table}

\begin{figure*}[!t]
\centering
\includegraphics[width=.88\textwidth]{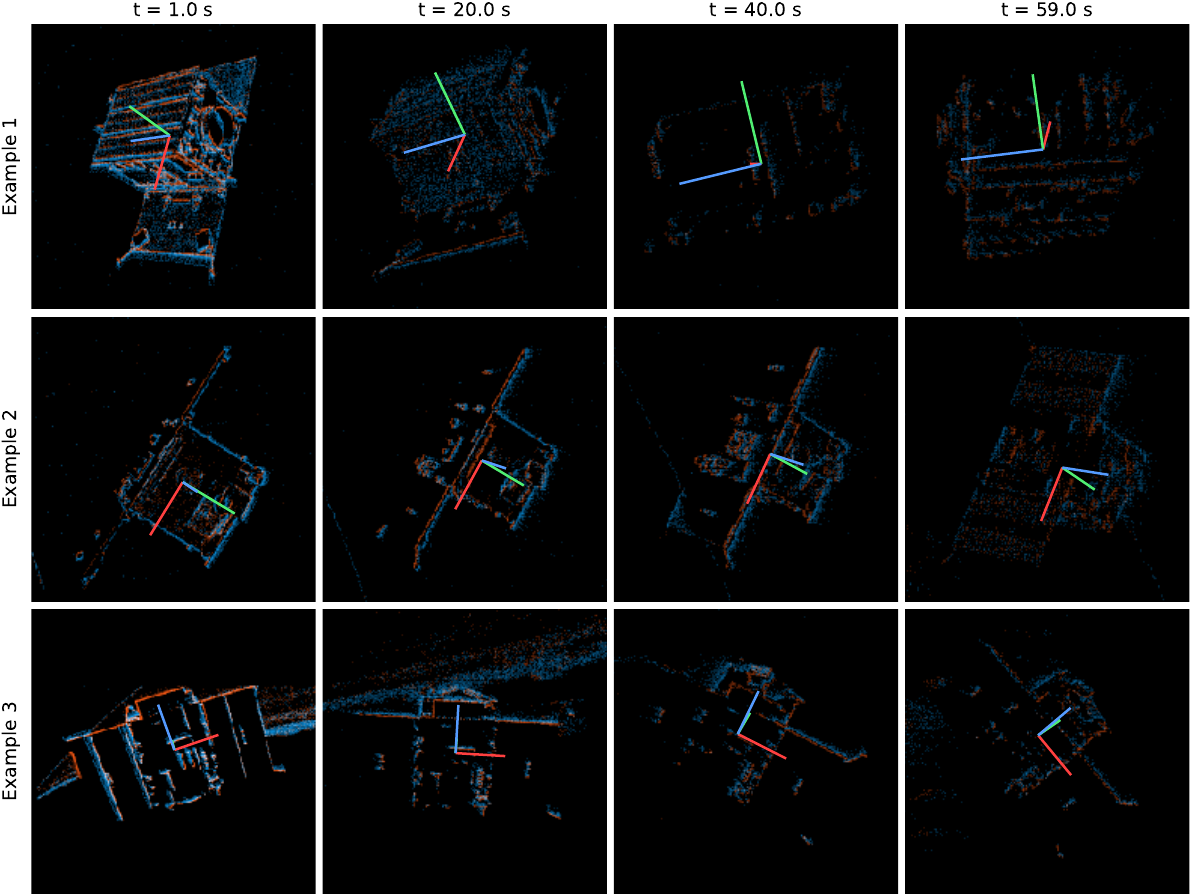}
\caption{Per-frame observations $T_k^{\mathrm{obs}}$ after ECR and before MQKS temporal optimization. Rows show three example sequences, and columns show four timestamps. Orange and blue pixels represent positive- and negative-polarity events, respectively. Red, green, and blue line segments denote the projected object-frame $X$, $Y$, and $Z$ axes, with a common physical length across all panels.}
\label{fig:local-refinement}
\end{figure*}

\subsubsection{Three-Branch Interaction and Reliability-Guided Recovery}

Table~\ref{tab:frontend-design} tests localization changes first, then correspondence weighting. STA is retained throughout, and ECR and MQKS are disabled. Starting from the dual-polarity baseline, recent-event timing, cross-branch attention, and gating are introduced in sequence. These configurations share the coordinate decoder, EPnP-RANSAC initialization, and robust refinement, with relative precision set to the identity. The final three rows hold the full interaction network's heatmaps and coordinates fixed. They compare uniform geometric weights with learned reliability trained using localization-error likelihood alone and with the additional pose-propagation proxy.

The localization comparison ends at a combined score of 0.0729 under uniform geometric weighting. Starting from 0.0772, recent-event timing gives 0.0762 and cross-branch attention gives 0.0739; gating completes the full interaction network. Its heatmaps and coordinates are then fixed for the correspondence-weighting comparison. NLL-based weighting yields 0.0716, with the pose-propagation proxy bringing the score to 0.0697.

\begin{table}[!t]
\centering
\caption{Progressive RPI ablation on the validation set.}
\label{tab:frontend-design}
\small
\setlength{\tabcolsep}{4pt}
\renewcommand{\arraystretch}{1.08}
\begin{tabularx}{\columnwidth}{Xrrr}
\toprule
Configuration & $E_t$ & $E_R$ & $E_{\mathrm{pose}}$ \\
\midrule
Dual-polarity baseline & 0.0208 & 0.0564 & 0.0772 \\
\quad + Recent-event time & 0.0204 & 0.0558 & 0.0762 \\
\quad + Cross-branch attention & 0.0200 & 0.0539 & 0.0739 \\
\quad + Channel gating & 0.0197 & 0.0532 & 0.0729 \\
\midrule
\quad + Reliability (NLL) & 0.0195 & 0.0521 & 0.0716 \\
\quad + Pose proxy (full RPI) & \textbf{0.0191} & \textbf{0.0506} & \textbf{0.0697} \\
\bottomrule
\end{tabularx}
\end{table}

\subsubsection{Contributions of Multi-Source Quality}

Table~\ref{tab:quality-sources} separates the effect of motion smoothing from the information used to weight observations. Every row receives the same post-ECR observations $T_k^{\mathrm{obs}}$; the first row reports them without trajectory optimization. Constant-weight optimization establishes the contribution of temporal constraints alone. Heatmap quality and geometry quality use their respective descriptor subsets; full MQKS combines them. The geometry subset comprises reprojection consistency and geometric conditioning. All quality mappings follow the same calibration procedure, and all smoothed configurations retain the same motion factors, initialization, solver, and optimization budget.

Most of the improvement over per-frame estimates comes from motion smoothing: constant-weight optimization reduces the combined score from 0.0674 to 0.0549. Quality weighting changes the influence of the observations within that optimization. Full MQKS reaches 0.0514, an additional reduction of approximately 6.4\% relative to constant weighting; using heatmap quality or geometry quality alone yields 0.0536 and 0.0522, respectively.

\begin{table}[!t]
\centering
\caption{Quality-source ablation on identical validation observations after ECR.}
\label{tab:quality-sources}
\small
\begin{tabularx}{\columnwidth}{Xrrr}
\toprule
Configuration & $E_t$ & $E_R$ & $E_{\mathrm{pose}}$ \\
\midrule
Per-frame ECR poses & 0.0185 & 0.0489 & 0.0674 \\
Constant-weight optimization & 0.0158 & 0.0391 & 0.0549 \\
Heatmap-quality optimization & 0.0156 & 0.0380 & 0.0536 \\
Geometry-quality optimization & 0.0152 & 0.0370 & 0.0522 \\
MQKS & \textbf{0.0150} & \textbf{0.0364} & \textbf{0.0514} \\
\bottomrule
\end{tabularx}
\end{table}

\begin{figure}[!t]
\centering
\includegraphics[width=\columnwidth]{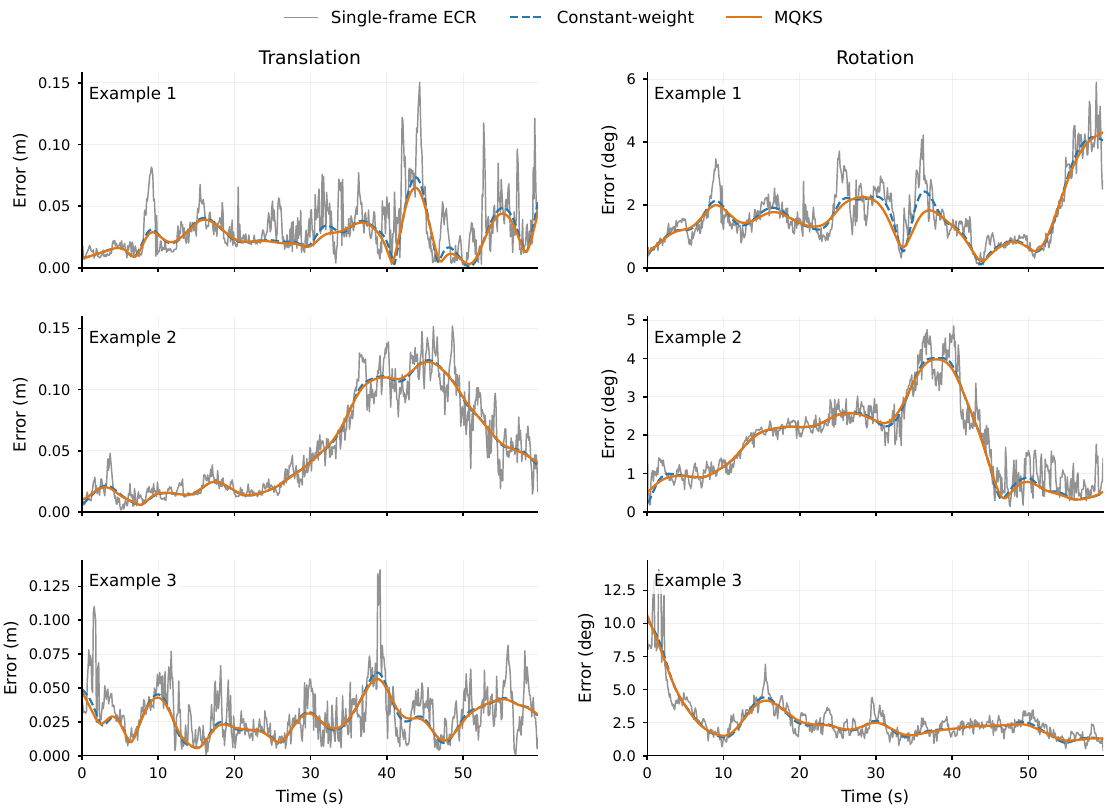}
\caption{Continuous-sequence errors for the three sequences in Fig.~\ref{fig:pose-overlays} (seed 2026). Left: absolute translation error in meters. Right: rotation error in degrees. Gray: per-frame ECR observations (``Single-frame ECR''); blue dashed: constant-weight optimization; orange: MQKS. All 599 timestamps, including error peaks, are shown without display smoothing.}
\label{fig:sequence-errors}
\end{figure}

\subsection{Temporal Window Analysis}

The window-length study in Table~\ref{tab:window-length} varies the input context over 1, 3, 5, 7, and 9 windows under matched training budgets. Recovery, refinement, and smoothing settings remain fixed. Every row evaluates per-frame output followed by trajectory optimization and reports translation, rotation, and combined error together with runtime.

Five windows give the lowest combined score among the tested configurations. Increasing context from one to five reduces the score from 0.0677 to 0.0514 and raises runtime from 36.8 to 52.7\,ms/frame. Longer contexts offer no further reduction in combined error: seven windows yield 0.0515 at 61.3\,ms/frame, with slightly better translation but worse rotation; nine yield 0.0523 at 70.2\,ms/frame.

\begin{table}[!t]
\centering
\caption{Validation accuracy and processing cost versus input-window length.}
\label{tab:window-length}
\small
\setlength{\tabcolsep}{5pt}
\begin{tabularx}{\columnwidth}{Xrrrr}
\toprule
Windows & $E_t$ & $E_R$ & $E_{\mathrm{pose}}$ & ms/frame \\
\midrule
1 & 0.0194 & 0.0483 & 0.0677 & 36.8 \\
3 & 0.0162 & 0.0392 & 0.0554 & 44.6 \\
5 & 0.0150 & 0.0364 & 0.0514 & 52.7 \\
7 & 0.0149 & 0.0366 & 0.0515 & 61.3 \\
9 & 0.0151 & 0.0372 & 0.0523 & 70.2 \\
\bottomrule
\end{tabularx}
\end{table}

\subsection{Qualitative Results}

Figure~\ref{fig:pose-overlays} shows the full RPI--ECR--MQKS output and ground truth on three preselected validation sequences. Their projected CAD bounding boxes share the same image coordinates, providing a reference for position, scale, and orientation at four fixed timestamps per sequence. A fixed display margin around the original ROI makes the projected box visible; network inputs and pose estimates use the original ROI.

Figure~\ref{fig:local-refinement} displays $T_k^{\mathrm{obs}}$ before MQKS. The estimated object coordinate axes are projected onto the corresponding event ROIs for the same three sequences and fixed timestamps as Fig.~\ref{fig:pose-overlays}. Together, the two figures distinguish the per-frame ECR observation from the final trajectory estimate.

Figure~\ref{fig:sequence-errors} uses the same three sequences as Fig.~\ref{fig:pose-overlays} and compares the per-frame ECR observations, constant-weight optimization, and MQKS over all 599 timestamps. The two temporal variants receive identical observations and use the same motion and robust-loss settings; they differ in observation-quality weighting. Translation error is $\|\widehat{\mathbf t}_k-\mathbf t_k^*\|_2$ in meters, and rotation error is the geodesic angle between the estimated and ground-truth rotations in degrees. Unlike the table scores, these curves use absolute translation error and unthresholded rotation error.

Both temporal variants suppress rapid fluctuations, but the sustained error near the middle of Example~2 remains. MQKS gives modestly lower mean rotation error in these examples without uniformly reducing every instantaneous error peak relative to constant weighting. These sequence-level observations accompany the aggregate accuracy evaluation over the complete validation set.

\section{Conclusion}

The framework carries measurement reliability from keypoint correspondences to the final pose trajectory, with different quantities governing the two levels. Robust pose recovery uses relative point precision learned from local polarity--time responses and ordered CAD geometry. Trajectory weights describe the accepted observations and are computed separately for translation and rotation. Between these steps, event--CAD agreement supplies additional edge evidence, while geometric safeguards control candidate acceptance.

On the official SPADES challenge test set, the complete system reaches a combined score of 0.0474, 5.4\% below PASAN+IRIS. Validation ablations reduce the score from 0.0772 to 0.0514; on identical observations, quality weighting accounts for a further 6.4\% reduction over constant-weight optimization. Five input windows give the lowest combined score among the tested lengths at a reported cost of 52.7\,ms/frame. The continuous-sequence results show the remaining difficulty: smoothing suppresses rapid fluctuations, while sustained deviations from ground truth persist. Persistent observation bias and temporal look-ahead remain targets for further work.

\bibliographystyle{IEEEtran}
\bibliography{references}

\begin{IEEEbiography}[{\includegraphics[width=1in,height=1.25in,clip,keepaspectratio]{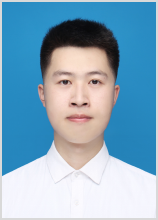}}]{Yonglong Zhang}
received his bachelor's degree in Mechatronic Engineering from Harbin Institute of Technology in 2024. He is currently pursuing the master's degree with the School of Mechatronics Engineering, Harbin Institute of Technology. His research interests include robot pose estimation and robotic arm manipulation.
\end{IEEEbiography}

\begin{IEEEbiography}[{\includegraphics[width=1in,height=1.25in,clip,keepaspectratio]{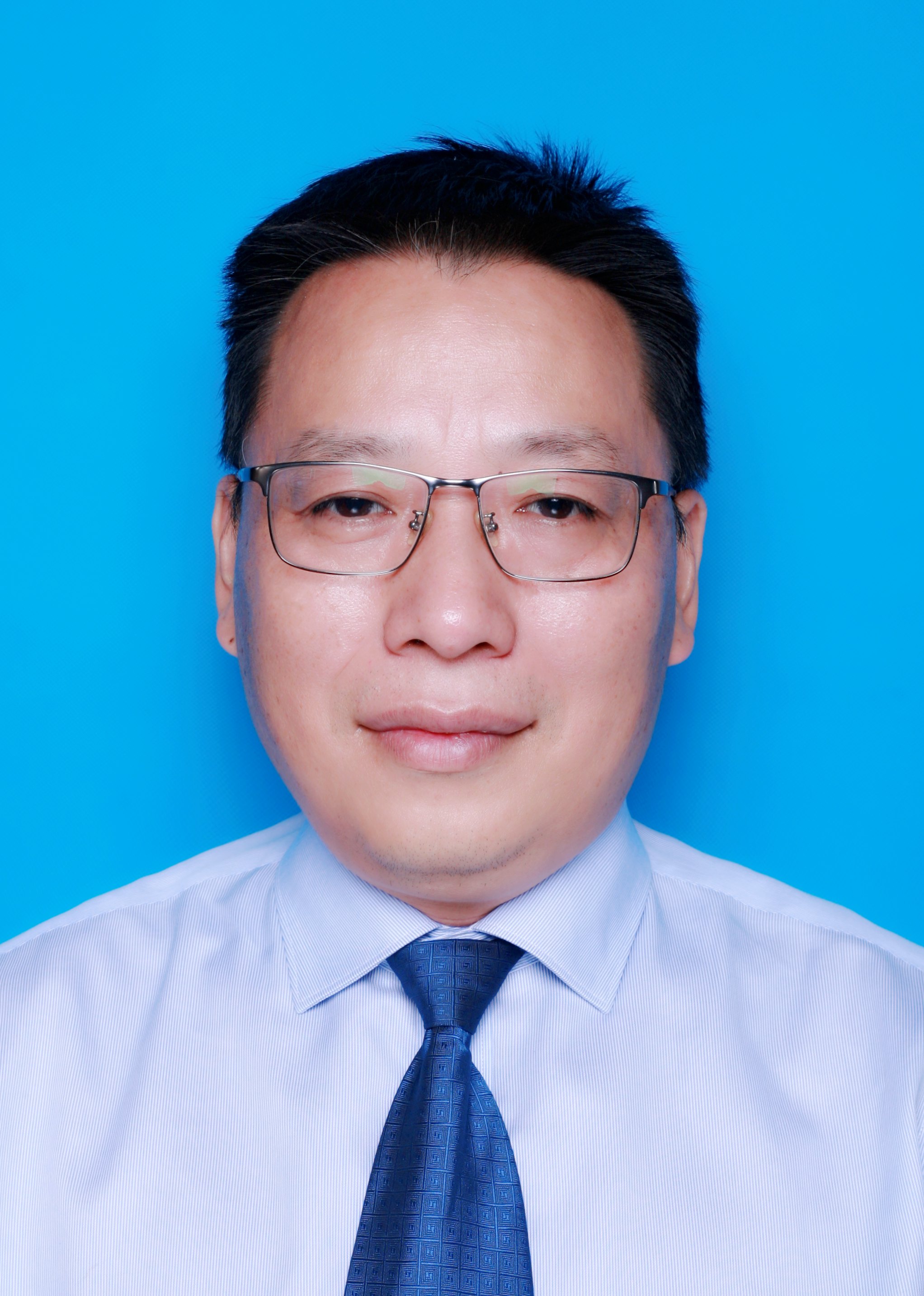}}]{Zongwu Xie}
received the B.S. degree in electrical engineering and automation from Harbin University of Science and Technology, Harbin, China, in 1996, and the M.S. and Ph.D. degrees in mechanical engineering from Harbin Institute of Technology, Harbin, in 2000 and 2003, respectively.
\end{IEEEbiography}

\begin{IEEEbiography}[{\includegraphics[width=1in,height=1.25in,clip,keepaspectratio]{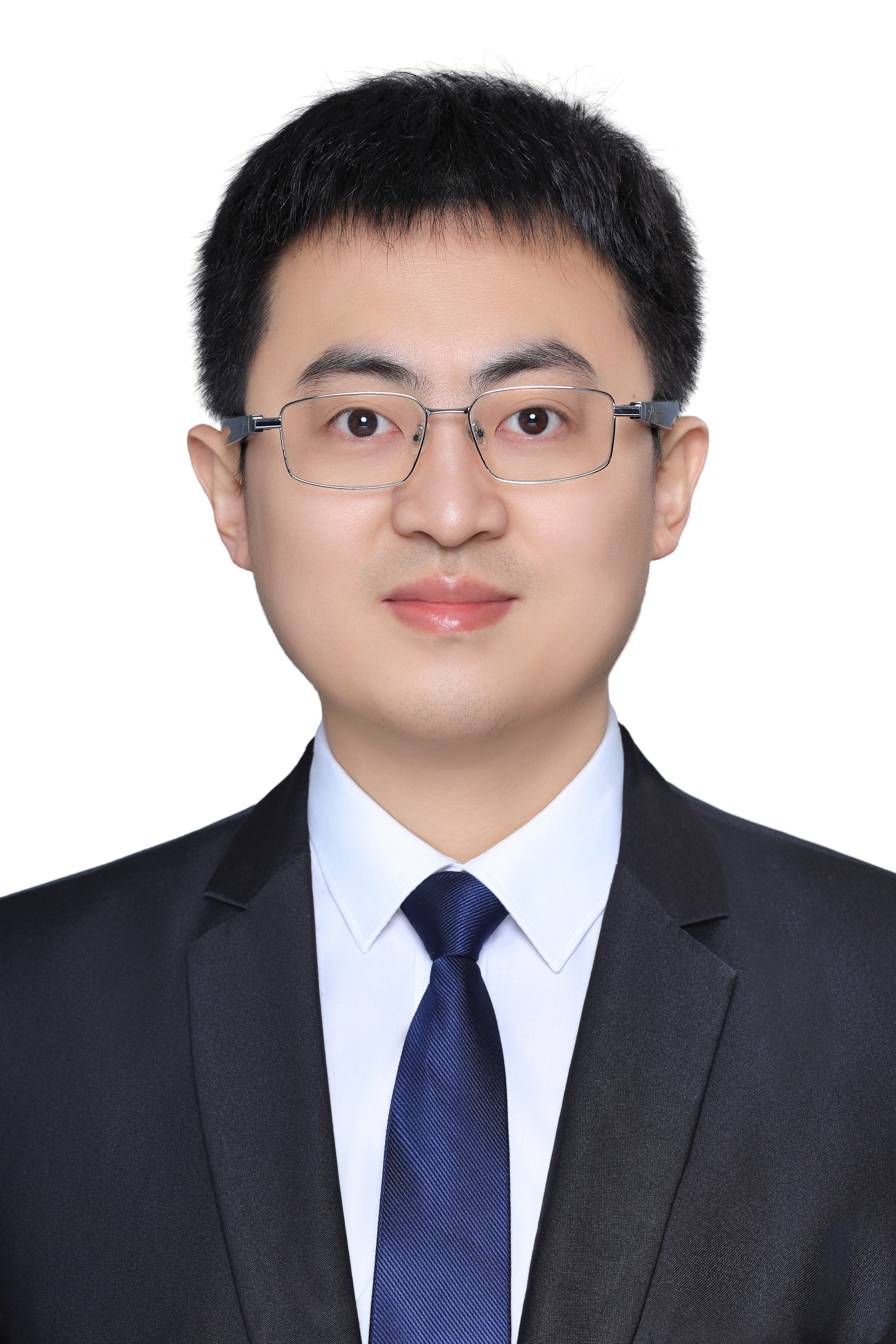}}]{Yang Liu}
was born in Handan, Hebei, China, in 1990. He received the B.S., M.S., and Ph.D. degrees from Harbin Institute of Technology, Harbin, in 2013, 2015, and 2020, respectively. He is currently an associate professor at Harbin Institute of Technology and serves as the deputy secretary of the Youth League Committee. His research interests include space robotics, humanoid robots, and computer vision. He has published 35 scientific papers and two textbooks and holds 16 invention patents. He has been selected for the Youth Talent Support Program and honored with titles including Heilongjiang Province Outstanding Youth.
\end{IEEEbiography}
\vfill
\end{document}